\documentclass{article}
\usepackage{spconf,amsmath,graphicx}
\usepackage{subfigure}
\usepackage{multirow}
\usepackage{array}
\usepackage{color}
\usepackage{amssymb}
\usepackage{tabu}
\usepackage{bibspacing}
\usepackage{cite}

\title{Weakly Supervised Lesion Co-segmentation on CT Scans}

 \name{Vatsal Agarwal$^{\ \star  \mathsection  \dagger}$ \quad Youbao Tang$^{\ \star  \dagger }$ \quad Jing Xiao$^{\ \ddagger}$ \quad Ronald M. Summers$^{\ \dagger}$ \thanks{$^{\star}$ equal contribution, $^{\mathsection}$ work done during internship in NIH while pursuing undergraduate degree at University of Maryland.}}

 \address{$^{\dagger}$ Imaging Biomarkers and Computer-Aided Diagnosis Laboratory, Radiology and Imaging Sciences, \\ National Institutes of Health Clinical Center, Bethesda, MD 20892, USA \\
     $^{\ddagger}$Ping An Insurance Company of China, Shenzhen 510852, China}

\begin{document}
%
\maketitle

\begin{abstract}

Lesion segmentation in medical imaging serves as an effective tool for assessing tumor sizes and monitoring changes in growth. However, not only is manual lesion segmentation time-consuming, but it is also expensive and requires expert radiologist knowledge. Therefore many hospitals rely on a loose substitute called response evaluation criteria in solid tumors (RECIST). Although these annotations are far from precise, they are widely used throughout hospitals and are found in their picture archiving and communication systems (PACS). Therefore, these annotations have the potential to serve as a robust yet challenging means of weak supervision for training full lesion segmentation models. In this work, we propose a weakly-supervised co-segmentation model that first generates pseudo-masks from the RECIST slices and uses these as training labels for an attention-based convolutional neural network capable of segmenting common lesions from a pair of CT scans. To validate and test the model, we utilize the DeepLesion dataset, an extensive CT-scan lesion dataset that contains 32,735 PACS bookmarked images. Extensive experimental results demonstrate the efficacy of our co-segmentation approach for lesion segmentation with a mean Dice coefficient of 90.3\%.  

\end{abstract}

\section{Introduction}
\label{sec:intro}
Quantitative assessment of changes in lesion/tumor growth remains a challenging problem within medical imaging and precision medicine. In the clinical setting, radiologists must manually assert tumor size using response evaluation criteria in solid tumors (RECIST). This requires them to find and annotate areas which roughly have the greatest cross area. Such a process is subjective and results in a lack of consistency that is needed to make proper diagnoses. Thus, the ability to directly compute lesion segmentations would be crucial in planning treatments as well as track and record lesion growth rates. Since manual segmentation is laborious in time and cost, RECIST is used as the default clinical standard throughout hospitals. This has resulted in a substantial number of RECIST measurements being stored in hospitals' picture archiving and communications systems (PACS) that correspond to patient CT scans. Inspired by the successes of deep learning in recent years, many medical image analysis applications \cite{jin2018ct,tang2018semi,cai2018accurate,tang2018ct,tang2019xlsor,yan2019mulan,tang2019lesiondetection,tang2019ct,tang2019tuna,8759442,tang2019deep,spie2020} have utilized deep learning to improve their performance. Some of them have been developed for lesion detection \cite{yan2019mulan,tang2019lesiondetection}, segmentation \cite{cai2018accurate,tang2018ct} and RECIST estimation \cite{tang2018semi} using RECIST as supervision. In \cite{spie2020}, we are the first to leverage existing RECIST diameters as weak supervision for a segmentation method that employs an attention-based co-segmentation design to obtain final lesion masks. In this paper, we build upon [6] by modifying our architecture to obtain and preserve higher resolution feature maps and experimenting with different training methods and attention mechanisms to improve performance.

\begin{figure*}[t!]
	\begin{center}
		\includegraphics[width=1\linewidth]{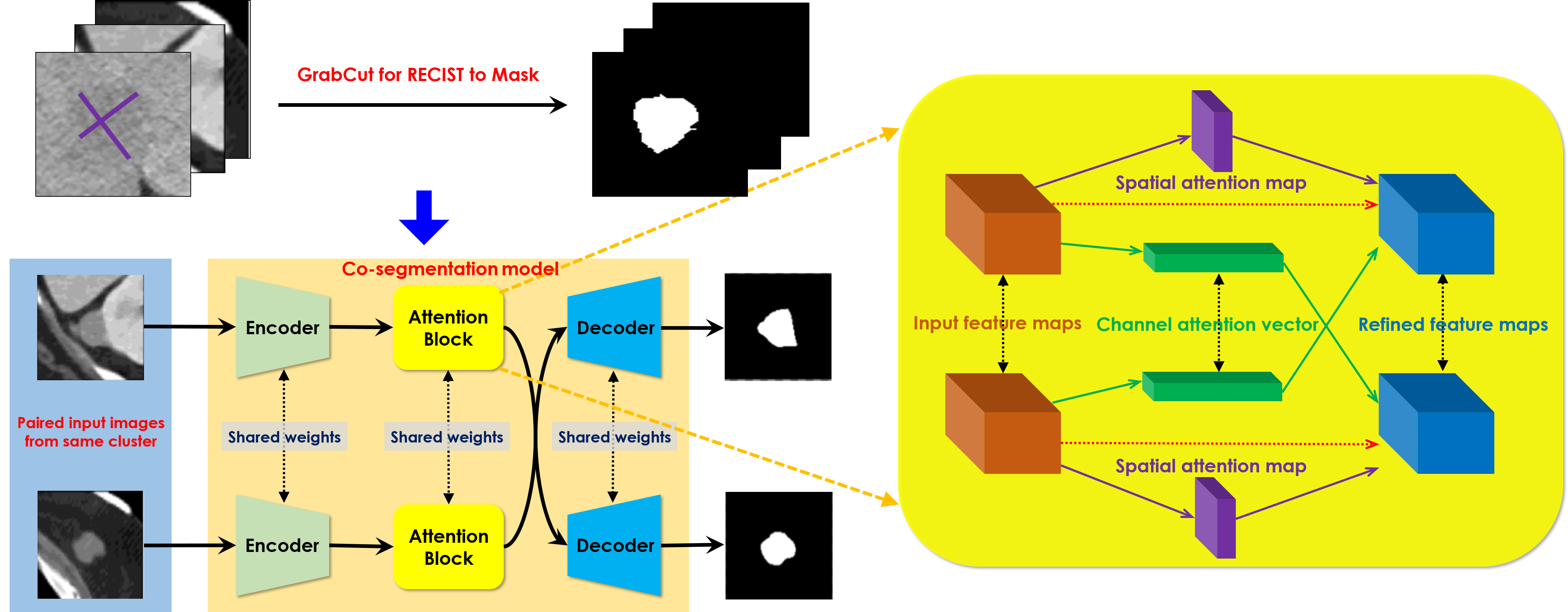}
	\end{center}
	\caption{The framework of the co-segmentation approach. For the attention block, we extract channel attention and spatial attention maps independently and apply them through element-wise multiplication.}
	\label{fig:model-pipeline}
\end{figure*}

In \cite{cai2018accurate}, GrabCut \cite{rother2004grabcut} is used to create initial lesion segmentations on RECIST slices for training a holistically nested network to generate the final masks. However, it has been shown that co-segmentation \cite{spie2020,rother2006cosegmentation}, which is the task of jointly segmenting common objects in a pair of images, can leverage similarities in appearance, background and semantic information to produce more accurate segmentations. Inspired by this intuition, this work leverages RECIST diameters as weak supervision for a convolutional neural network based  co-segmentation method that uses attention to obtain refined lesion masks. Due to significant variance between different lesions in size and appearance, this work proposes utilizing lesion embeddings to cluster the lesions into 200 classes prior to training the model.

\section{Our Lesion Co-Segmentation Method}
\label{sec:intro}

In this work, we present a weakly-supervised co-segmentation approach to generate 2D lesion segmentations. Our two-tiered approach is shown in Fig.~\ref{fig:model-pipeline}. First, initial lesion masks are generated using RECIST markers. We then train a robust attention-based co-segmentation model on these masks and use it to produce the final segmentations. RECIST diameters (indicated by the purple cross in Fig.~\ref{fig:model-pipeline}) act as the weakly-supervised training data. The details are described below.

\subsection{Pseudo-mask Generation} 
The NIH DeepLesion dataset \cite{yan2018deeplesion} only provides the RECIST annotation for each lesion region due to the high cost of manual generation of full lesion segmentation masks. Thus, it is necessary to create pseudo-masks from the RECIST annotations using unsupervised learning methods that can then be used to train a supervised model. A popular method for this task is GrabCut, which is first initialized with seeds for the image foreground and background regions and then uses iterative energy minimization to produce initial masks. Following \cite{cai2018accurate}, we leverage RECIST-slice information and compute the seeds using GrabCut from the RECIST markers to obtain initial lesion masks(as shown in the top part of Fig.~\ref{fig:model-pipeline}). We then use these masks as training data for the co-segmentation model. 

\subsection{Lesion Co-segmentation}
\label{sec:lesion co-segmentation}
After the initial lesion masks are generated, we train the co-segmentation model for lesion segmentation. The bottom part of Fig.~\ref{fig:model-pipeline} shows a diagram of how the model is trained using the initial masks. We adopt the co-segmentation model from \cite{chen2018semantic}. The model consists of a Siamese encoder and decoder and utilizes an attention module in the bottleneck layers. Using a pair of images as input, the model produces a mask for each lesion that is then fed to a densely-connected conditional random field \cite{krahenbuhl2011efficient} (DCRF) to acquire the final masks.  

A channel-attention module and a spatial-attention module comprise our attention mechanism. The right side of Fig.~\ref{fig:model-pipeline} shows the channel-spatial attention mechanism (CSA). We take the feature maps from the encoder network and apply the attention modules to emphasize common semantic information and suppress the rest. By preserving channels that contribute the most to both feature maps, the channel-attention module enhances shared information. The spatial-attention module captures spatial details in the feature maps. 

With regards to encoder design, we employ a pre-trained ResNet-101 \cite{he2016deep} and add atrous convolutions to the final two residual blocks \cite{chen2017deeplab}, creating feature maps with an output stride of 8 (DRN-101). Thus, we can attain higher-resolution feature maps that contain greater context. We also examine using a multi-grid strategy for the encoder (DRN-MG-101) in order to extract feature maps at different scales. For our decoder, we experiment with two models. The first is adopted from \cite{chen2018semantic} with slight modifications (i.e. less convolutional layers, D1), while the second is adopted from \cite{chen2018encoder} and takes in both the attended and lower level feature maps to create a more robust segmentation mask (D2). Please refer to \cite{chen2018encoder} for additional details. When using the improved decoder (D2), we refrain from using DCRF as it performs worse.

\begin{table*}[t!]
\caption{The performance of lesion segmentation using a variety of different strategies. For the co-segmentation models, the original decoder network is used.The means and standard deviations are reported for all strategies. $\uparrow$: the larger the better. $\downarrow$: the smaller the better.} 
\label{tab:training_coseg_performance}
\small
\begin{center}   
\begin{tabular}{|@{}*{1}{m{3.2cm}<{\centering}@{} |@{}}*{3}{m{1.1cm}<{\centering} @{}|@{}}*{5}{m{2.2cm}<{\centering}@{}|@{}}}

\hline
\rule[-1ex]{0pt}{3.5ex} \textbf{Model} & \textbf{MG} & \textbf{Adam} & \textbf{SGD} & \textbf{Rec.}$\uparrow$  & \textbf{Prec.}$\uparrow$  & \textbf{Dice}$\uparrow$  & \textbf{AVD}$\downarrow$ & \textbf{VS}$\uparrow$   \\
\hline
\rule[-1ex]{0pt}{3.5ex}  DRN-101 FCN-32 & & & & 
0.917 $\pm$ 0.13  & 0.882 $\pm$ 0.11  & 0.893 $\pm$ 0.10  & 0.396 $\pm$ 1.39  & 0.938 $\pm$ 0.07  \\
\rule[-1ex]{0pt}{3.5ex}  DRN-101 CSA + D1&  & \checkmark &  &
{0.915 $\pm$ 0.11}  & 0.895 $\pm$ 0.10  & {0.898 $\pm$ 0.10} & {0.349 $\pm$ 1.20}  & {0.942 $\pm$ 0.08} \\
\rule[-1ex]{0pt}{3.5ex}  DRN-101 CSA + D1&  &  & \checkmark &
\textbf{0.918 $\pm$ 0.11}  & 0.890 $\pm$ 0.10  & {0.898 $\pm$ 0.10} & {0.357 $\pm$ 1.18}  & {0.942 $\pm$ 0.07} \\
\rule[-1ex]{0pt}{3.5ex}  DRN-101 CSA + D1& \checkmark & \checkmark &  &
0.899 $\pm$ 0.12  & \textbf{0.910 $\pm$ 0.10}  & 0.898 $\pm$ 0.10  & 0.355 $\pm$ 1.22  & 0.943 $\pm$ 0.08  \\
\rule[-1ex]{0pt}{3.5ex}  DRN-101 CSA + D1& \checkmark & & \checkmark &  
0.907 $\pm$ 0.11  & 0.906 $\pm$ 0.10  & \textbf{0.901 $\pm$ 0.09}  & \textbf{0.342 $\pm$ 1.21} & \textbf{0.946 $\pm$ 0.07}  \\
\hline
\end{tabular}
\end{center}
\end{table*}

\subsection{Attention Modules}
\label{sec:attention modules}
Our base mechanism takes inspiration from \cite{chen2018semantic} and uses a Squeeze and Excitation network \cite{hu2018squeeze} to obtain the channel attention vectors (CA = SE) and calculate the mean value at each spatial location over all feature maps to obtain the spatial attention maps (SA = MSA). In order to improve upon this, we experiment with three methods for obtaining channel and spatial attention maps. Introduced in \cite{chen2017deeplab}, atrous spatial pyramid pooling can be used to provide robust multi-scale information. Due to its ability to capture long-range contextual information through a series of varied atrous convolutions, we investigate its performance as a spatial attention mechanism (SA = ASPP). Our second attention mechanism focuses on improvements to channel attention. Following \cite{wang2019eca}, we modify the SENet architecture, and replace the fully-connected layer with a 1D convolutional layer with an adaptive kernel size to better capture cross-channel interactions (CA = ECA). The third attention mechanism is adopted from \cite{fu2019dual} and consists of a position attention module and channel attention module (CSA = DANet). Both modules utilize a similarity matrix to obtain a weighted sum of features across all positions and channels respectively along with the original features. Please refer to \cite{wang2019eca} and \cite{fu2019dual} respectively for additional details. 

\subsection{Implementation Details}
All models are implemented in PyTorch\cite{paszke2017automatic}. Training is conducted with a batch-size of 20 images for a total of two epochs of 12,000 iterations each. We use the Adam optimizer with a learning rate of $1\mathrm{e}{-5}$ and a weight decay of 0.0005 for L2 regularization and the SGD optimizer with a weight decay of 0.0005 and momentum of 0.99. The initial learning rate is 0.01 and we decay the learning rate following a poly-learning rate policy where the initial learning rate is multiplied by $(1 - \frac{iter}{total_{iter}})^{0.9}$ after each iteration. Prior to training, we pad lesion-mask images before resizing them to 128x128 and then normalize them.

\begin{table*}[t!]
\caption{The performance of lesion segmentation with the two different decoders along with CRF post-processing trained with SGD poly policy.} 
\label{tab:decoder_coseg_performance}
\small
\begin{center}   
\begin{tabular}{|@{}*{1}{m{3.2cm}<{\centering}@{} |@{}}*{3}{m{1.1cm}<{\centering} @{}|@{}}*{5}{m{2.2cm}<{\centering}@{}|@{}}}

\hline
\rule[-1ex]{0pt}{3.5ex} \textbf{Model} & \textbf{D1} & \textbf{D2} & \textbf{DCRF} & \textbf{Rec.}$\uparrow$  & \textbf{Prec.}$\uparrow$  & \textbf{Dice}$\uparrow$  & \textbf{AVD}$\downarrow$ & \textbf{VS}$\uparrow$   \\
\hline
\rule[-1ex]{0pt}{3.5ex}  DRN-MG- 101 CSA & \checkmark &  &  &
\textbf{0.921 $\pm$ 0.11}  & 0.890 $\pm$ 0.10  & {0.899 $\pm$ 0.09} & {0.350 $\pm$ 1.18}  & {0.944 $\pm$ 0.07} \\
\rule[-1ex]{0pt}{3.5ex}  DRN-MG- 101 CSA & \checkmark &  & \checkmark &
0.907 $\pm$ 0.11  & {0.906 $\pm$ 0.10}  & {0.901 $\pm$ 0.09}  & {0.342 $\pm$ 1.21} & \textbf{0.946 $\pm$ 0.07}  \\
\rule[-1ex]{0pt}{3.5ex}  DRN-MG- 101 CSA & & \checkmark &  &
0.920 $\pm$ 0.11  & {0.898 $\pm$ 0.10}  & \textbf{0.903 $\pm$ 0.09}  & \textbf{0.321 $\pm$ 1.15}  & 0.943 $\pm$ 0.07  \\
\rule[-1ex]{0pt}{3.5ex}  DRN-MG- 101 CSA & & \checkmark & \checkmark &
0.905 $\pm$ 0.11  & \textbf{0.908 $\pm$ 0.09}  & 0.901 $\pm$ 0.09  & 0.326 $\pm$ 1.16  & 0.945 $\pm$ 0.07  \\
\hline
\end{tabular}
\end{center}
\vspace{-0.5\baselineskip}
\end{table*}

\begin{table*}[t!]
\caption{The performance of lesion segmentation with different attention mechanisms trained with SGD poly-learning policy and the improved decoder architecture.}  
\label{tab:attentions_coseg_performance}
\small
\begin{center}       

\begin{tabular}{|@{}*{1}{m{3cm}<{\centering}@{} |@{}}*{4}{m{1cm}<{\centering} @{} |@{}}*{1}{m{1cm}<{\centering} @{} | @{}}*{5}{m{1.9cm}<{\centering}@{}|@{}}}

\hline
\rule[-1ex]{0pt}{3.5ex} \textbf{Model} & 
\textbf{ CA = SE} & \textbf{ CA = ECA} & \textbf{ SA = MSA} & \textbf{ SA = ASPP} & \textbf{  CSA = DANet} &
\textbf{Rec.}$\uparrow$  & \textbf{Prec.}$\uparrow$  & \textbf{Dice}$\uparrow$  & \textbf{AVD}$\downarrow$ & \textbf{VS}$\uparrow$   \\
\hline

\rule[-1ex]{0pt}{3.5ex}   DRN-MG- 101 + D2& \checkmark &  & \checkmark &  &  & 
0.920 $\pm$ 0.11  & \textbf{0.898 $\pm$ 0.10}  & \textbf{0.903 $\pm$ 0.09}  & \textbf{0.321 $\pm$ 1.15}  & 0.943 $\pm$ 0.07  \\
\rule[-1ex]{0pt}{3.5ex}   DRN-MG- 101 + D2 & \checkmark &  &  & \checkmark &  &  
0.924 $\pm$ 0.11  & 0.892 $\pm$ 0.10  & 0.902 $\pm$ 0.09  & 0.335 $\pm$ 1.16  & 0.941 $\pm$ 0.07  \\
\rule[-1ex]{0pt}{3.5ex}   DRN-MG- 101 + D2 &  & \checkmark & \checkmark &  &  & 
\textbf{0.926 $\pm$ 0.11}  & 0.891 $\pm$ 0.10  & 0.902 $\pm$ 0.09  & 0.331 $\pm$ 1.16  & 0.941 $\pm$ 0.07  \\
\rule[-1ex]{0pt}{3.5ex}  DRN-MG- 101 + D2 & & \checkmark & & \checkmark &  & 
0.921 $\pm$ 0.11  & 0.895 $\pm$ 0.10  & 0.902 $\pm$ 0.09  & 0.338 $\pm$ 1.17  & 0.943 $\pm$ 0.07  \\
\rule[-1ex]{0pt}{3.5ex}   DRN-MG- 101 + D2 & & & & & \checkmark &
0.921 $\pm$ 0.10  & 0.894 $\pm$ 0.10  & 0.902 $\pm$ 0.08  & 0.333 $\pm$ 1.15  & \textbf{0.944 $\pm$ 0.07}  \\
\hline
\end{tabular}
\end{center}
\end{table*}

\begin{figure*} [ht]
    \begin{center}
        \begin{tabular}{c} 
            \includegraphics[width=0.99\linewidth]{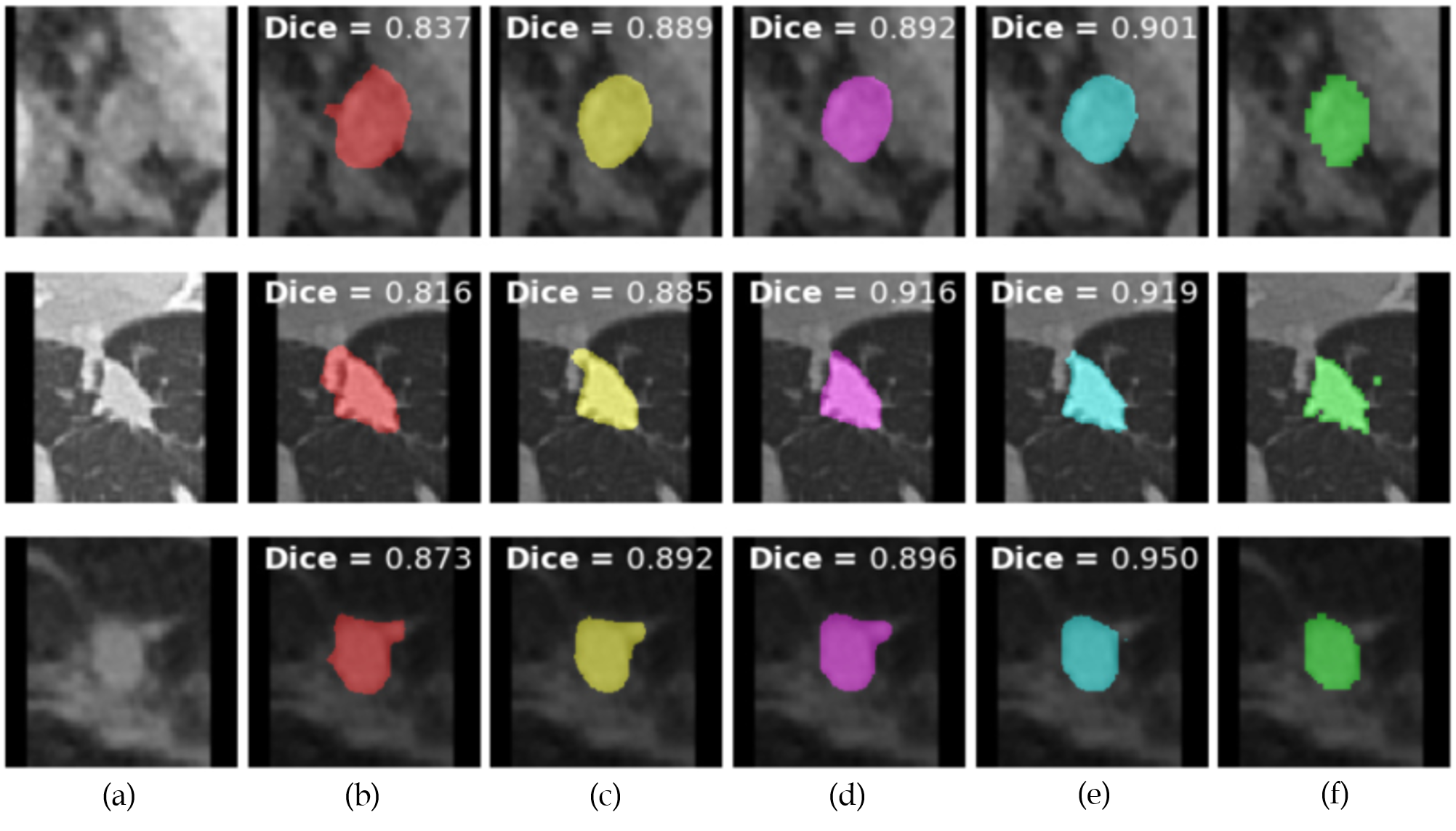}
        \end{tabular}
   \end{center}
\caption[results] 
{  Three lesion segmentation results using (b) FCN-32 with DRN-101 backbone, (c) DRN-MG-101  CSA + D1 with Adam training, (d) DRN-MG-101 CSA + D1 with poly training strategy, (e) DRN-MG-101 CSA + D2 with poly training strategy. (a) is the input image, while (f) is the ground truth.}
\label{fig:results}
\end{figure*} 

\section{Experiments}
\subsection{Dataset and Evaluation Criterion}
The NIH DeepLesion dataset \cite{yan2018deeplesion} is used for performance evaluation and is composed of $32,735$ PACS CT lesion images annotated with RECIST long and short diameters. These are derived from $10,594$ studies of $4,459$ patients. Following the clustering of the lesions into 200 classes using \cite{yan2018deep}, we split the dataset into 80\% training, 10\% validation and 10\% testing sets using stratified sampling. We pair the images from each cluster to build the co-segmentation dataset, resulting in 270,470 pairs for training, 28,136 pairs for validation and 3,866 pairs for testing. We hold out 1,000 manually annotated segmentations for evaluation and report the recall, precision, Dice similarity coefficient, averaged Hausdorff distance (AVD), and volumetric similarity (VS) for quantitative metrics, which are calculated pixel-wisely by a publicly available segmentation evaluation tool \cite{taha2015metrics}.

\subsection{Results and Analyses}

We first train a fully-convolutional network (FCN) \cite{long2015fully} with a dilated ResNet-101 backbone on the initial lesion masks to determine the efficacy of our co-segmentation approach. We then examine methods to improve feature extraction by using a multi-grid strategy and analyze various mechanisms to enhance our attention models' ability to capture inter-channel and inter-spatial information. The default attention mechanism is when CA = SE and SA = MSA and is denoted as CSA. In order to create robust segmentations from the attended feature maps, we also experiment with two previously mentioned decoder structures. Additionally, we explore how the SGD optimizer compares to the Adam optimizer for this task. 

\textbf{Quantitative Comparisons}: First, it can be seen from Table~\ref{tab:training_coseg_performance} that using co-segmentation with CSA improves upon the baseline DRN-101-backbone FCN with a 0.5\% increase in Dice score, thus validating our approach. Furthermore, we find that training using the SGD poly-learning policy enhances performance compared to using the Adam optimizer. The SGD optimized multi-grid model is seen to perform much better than its Adam-trained counterpart, demonstrating that although both optimizers obtain low training errors, adaptive methods may generalize worse on new data. With regards to encoder strategies, we find that a multi-grid strategy (DRN-MG-101) with dilation rates of 2, 4, and 8 improves performance for both training strategies, especially with the poly training policy where a 0.3\% increase in Dice score is noted for the SGD trained model. This indicates that extracting features at different scales is key to creating more robust segmentation masks. Additionally, the results in Table~\ref{tab:decoder_coseg_performance} validate that using the improved decoder with lower level features leads to more refined segmentations. For both decoders, the DCRF post-processing makes significant gains in precision in exchange for a sharp reduction in recall. 

We experiment with three different attention mechanisms to obtain more precise attended features for improved segmentations. The results are shown in Table~\ref{tab:attentions_coseg_performance}. First, it can be observed that the model using CSA has the highest performance with a Dice score of 90.3\%. Comparatively, minor decreases in performance are found with the introduced attention mechanisms. This could be due to better generalization capability as there are less parameters to fit with this attention model and thus there is less overfitting on the training set. The model also has a lower averaged Hausdorff distance of 0.321, revealing the strength of the attention mechanism in obtaining more precise boundary information. We find that using the efficient channel attention achieves the next-best performance, specifically with the ASPP spatial attention. Examining the results of the DANet channel-spatial attention, we do not find any significant increases in performance compared to using pooling-based methods for extracting feature attentions. 

\textbf {Qualitative Comparisons}:  Fig. 2 shows the qualitative results using the base DRN-101 FCN-32 along with three co-segmentation models using the different strategies. From this, it can be seen that the initial segmentations from the fully-convolutional network have significantly more pixels classified as false positives than those generated by the other models. This demonstrates the strength of the co-segmentation models in obtaining more precise boundaries by utilizing joint semantic information. The precision is further improved by using the multi-grid approach and can be attributed to extracting feature maps at different rates to produce more fine segmentation details. Using the improved decoder obtains predictions closest to the ground-truth, demonstrating that using lower level features for each image can better help the model to utilize independent semantic information and suppress possible interference in the attended feature maps. 

\section{Conclusions}
In this work, a weakly-supervised attention-based deep co-segmentation approach is proposed to obtain accurate lesion segmentations from RECIST measurements. GrabCut is first used to get initial lesion masks using RECIST diameters. Then, we leverage lesion embeddings to cluster lesions for co-segmentation. Our experimental results, both quantitative and qualitative, find that co-segmentation achieves promising results for weakly-supervised lesion segmentation along with an efficient training strategy. This model architecture has the potential to be employed in a slice-propagated manner to generate fully volumetric lesion segmentations. 

\vspace*{0.5\baselineskip}
\noindent\textbf{Acknowledgments.}
This research was supported by the Intramural Research Program of the National Institutes of Health Clinical Center and by the Ping An Insurance Company through a Cooperative Research and Development Agreement. We thank Nvidia for GPU card donation.

\bibliographystyle{IEEEbib}
\bibliography{refs}

\end{document}